\UseRawInputEncoding
\documentclass[10pt,twocolumn,letterpaper]{article}
\usepackage{cvpr}              % To produce the CAMERA-READY version
\usepackage[accsupp]{axessibility}
\usepackage{graphicx}
\usepackage{amsmath}
\usepackage{amssymb}
\usepackage{booktabs}
\usepackage{multirow}

\usepackage{capt-of}
\usepackage{times}
\usepackage{epsfig}
\usepackage{graphicx}
\usepackage{amsmath}
\usepackage{amssymb}
\usepackage{xcolor}
\usepackage{blindtext}
\usepackage{overpic}
\usepackage{transparent}
\usepackage{booktabs}

\usepackage[super]{nth}
\usepackage[format=plain,labelformat=simple,labelsep=period,font=small,skip=4pt,compatibility=false]{caption}
\usepackage[font=footnotesize,skip=2pt]{subcaption}
\usepackage{makecell}

\makeatletter
\newcommand\notsotiny{\@setfontsize\notsotiny\@vipt\@viipt}

\makeatother

\usepackage{etoolbox}
\usepackage[binary-units]{siunitx}
\robustify\bfseries
\DeclareSIUnit{\inch}{inch}

\newcolumntype{C}[1]{>{\centering\arraybackslash}p{#1}}

\usepackage{xspace}
\makeatletter
\DeclareRobustCommand\onedot{\futurelet\@let@token\@onedot}
\def\@onedot{\ifx\@let@token.\else.\null\fi\xspace}
\def\eg{\emph{e.g}\onedot} 
\def\ie{\emph{i.e}\onedot}

\makeatother

\usepackage{pifont}
\definecolor{lightgray}{rgb}{0.9, 0.9, 0.9}
\definecolor{lgray}{rgb}{0.66, 0.66, 0.66}
\usepackage[pagebackref,breaklinks,colorlinks]{hyperref}

\usepackage[capitalize]{cleveref}
\crefname{section}{Sec.}{Secs.}
\Crefname{section}{Section}{Sections}
\Crefname{table}{Table}{Tables}
\crefname{table}{Tab.}{Tabs.}

\begin{document}

%%%%%%%%% TITLE - PLEASE UPDATE
\title{High-fidelity Generalized Emotional Talking Face Generation \\
with Multi-modal Emotion Space Learning}

\author{Chao Xu$^1$
~ ~ Junwei Zhu$^2$
~ ~ Jiangning Zhang$^2$
~ ~ Yue Han$^1$
~ ~ Wenqing Chu$^2$ \\
~ ~ Ying Tai$^2$
~ ~ Chengjie Wang$^2$$^,$$^4$\thanks{Corresponding authors}
~ ~ Zhifeng Xie$^3$
~ ~ Yong Liu$^1$$^*$ \\
\normalsize $^1$ APRIL Lab, Zhejiang University ~ $^2$Youtu Lab, Tencent ~ $^3$Shanghai University，~ $^4$Shanghai Jiao Tong University \\
{\tt\small \{21832066, 22132041\}@zju.edu.cn, yongliu@iipc.zju.edu.cn, wqchu16@gmail.com} \\
{\tt\small \{junweizhu, vtzhang, yingtai, jasoncjwang\}@tencent.com, zhifeng\_xie@shu.edu.cn}
}

\maketitle
%%%%%%%%% ABSTRACT
\begin{abstract}
    Recently, emotional talking face generation has received considerable attention. However, existing methods only adopt one-hot coding, image, or audio as emotion conditions, thus lacking flexible control in practical applications and failing to handle unseen emotion styles due to limited semantics. They either ignore the one-shot setting or the quality of generated faces. In this paper, we propose a more flexible and generalized framework. Specifically, we supplement the emotion style in text prompts and use an Aligned Multi-modal Emotion encoder to embed the text, image, and audio emotion modality into a unified space, which inherits rich semantic prior from CLIP.
    Consequently, effective multi-modal emotion space learning helps our method support arbitrary emotion modality during testing and could generalize to unseen emotion styles. Besides, an Emotion-aware Audio-to-3DMM Convertor is proposed to connect the emotion condition and the audio sequence to structural representation. A followed style-based High-fidelity Emotional Face generator is designed to generate arbitrary high-resolution realistic identities. Our texture generator hierarchically learns flow fields and animated faces in a residual manner. Extensive experiments demonstrate the flexibility and generalization of our method in emotion control and the effectiveness of high-quality face synthesis.
\end{abstract}

%%%%%%%%% BODY TEXT
\section{Introduction}\label{sec:intro}
Talking face generation~\cite{guo2021ad, zhou2020makelttalk, yao2022dfa, song2022everybody} is the task of driving a static portrait with given audio. Recently, many works have tried to solve the challenges of maintaining lip movements synchronized with input speech contents and synthesizing natural facial motion simultaneously.
However, most researchers ignore a more challenging task, emotional audio-driven talking face generation, which is critical for creating vivid talking faces.

\begin{figure*}[t!]
	\centering
	\includegraphics[width=0.95\textwidth]{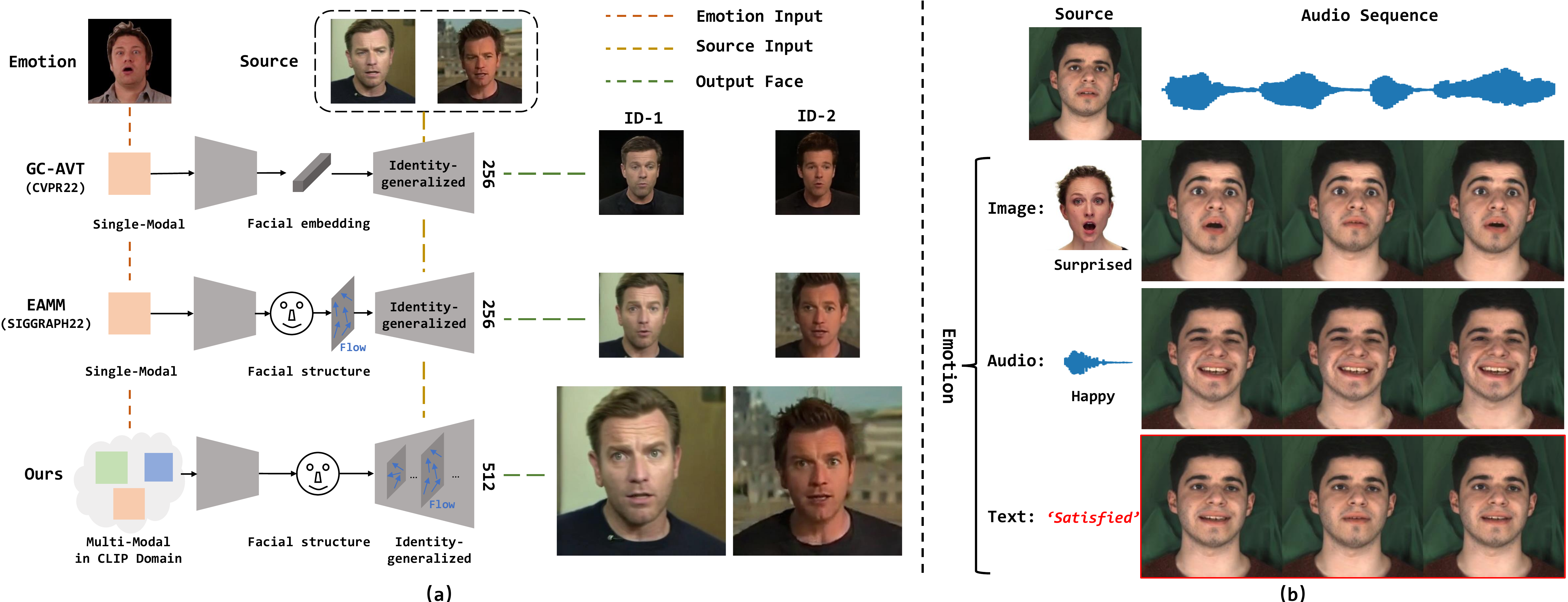}
	\caption{(a) An illustrative comparison of GC-AVT~\cite{liang2022expressive}, EAMM~\cite{ji2022eamm}, and our approach. First, our method supports \textit{multi-modal emotion cues} as input. As shown in (b), given a source face, an audio sequence, and diverse emotion conditions, our results fulfill synchronized lip movements with the speech content and emotional face with the desired style. Besides, benefiting from the effective multi-modal emotion space and rich semantics of CLIP, our method could generalize to \textit{unseen} style marked in \textcolor{red}{Red}. Second, the hierarchical style-based generator with coarse-to-fine facial deformation learning helps us generalize to unseen faces in high resolution and provides more \textit{realistic details and precise emotion} than GC-AVT and EAMM. Images are from the official attached results or released codes for fair comparisons. 
% 	Please zoom in for more details.
	}
	\label{fig:teaser}
\end{figure*}

Some works have achieved significant progress in solving the above task conditioned on emotion embedding. However, there are three continuously critical issues: 1) How to explore a more semantic emotion embedding to achieve better \textit{generalization for unseen emotions}. Early efforts~\cite{wang2020mead, ye2022dynamic, zhao2021emotion} adopt the one-hot vector to indicate emotion category, which could only cover the pre-defined label and lacks semantic cues. Subsequently, EVP~\cite{ji2021audio} disentangles emotion embedding from the audio, while GC-AVT~\cite{liang2022expressive} and EAMM~\cite{ji2022eamm} drive emotion by visual images. However, tailored audio- and image-based emotion encoders show limited semantics and also struggle to handle unseen emotion styles. 2) Could we construct \textit{multi-modal} emotion sources into a unified feature space to allow a more flexible and user-friendly emotion control. Existing methods only support one specific modality as the emotion condition, while the desired modality is usually not available in practical applications. 3) How to design a \textit{high-resolution identity-generalized generator}. Early works~\cite{wang2020mead, ji2021audio, ye2022dynamic} are in identity-specific design, while recent works~\cite{liang2022expressive, ji2022eamm} have started to enable one-shot emotional talking face generation. However, as shown in Fig.~\ref{fig:teaser}(a), GC-AVT and EAMM fail to produce high-resolution faces due to the inevitable information loss in face embedding and the challenge of estimating accurate high-resolution flow fields, respectively.

To address the aforementioned challenges, we first supplement the emotion styles in the text prompt inspired by the zero-shot CLIP-guided image manipulation~\cite{patashnik2021styleclip, tevet2022motionclip, wei2022hairclip}, which could inherit rich semantic knowledge and convenient interaction after being encoded. As shown in Fig.~\ref{fig:teaser}(b), unseen emotions, \eg., \textit{Satisfied}, could be flexibly specified using the text description and precisely reflected on the source face. Furthermore, to achieve alignment among multi-modal emotion features, we introduce an Aligned Multi-modal Emotion (AME) encoder to unify the text, image, and audio emotion modality into the same domain, thus supporting flexible emotion control by multi-modal inputs, as depicted in Fig.~\ref{fig:teaser}(b). In particular, the fixed CLIP text and image encoders are leveraged to extract their embedding and a learned CLIP audio encoder guided by several losses to find the proper emotion representation of the given audio sequence in CLIP space.

To this end, we follow the previous talking face generation methods~\cite{ren2021pirenderer} that rely on intermediate structural information such as 3DMM, and propose an Emotion-aware Audio-to-3DMM Convertor (EAC), to distill the rich emotional semantics from AME and project them to the facial structure. Specifically, we employ the Transformer~\cite{vaswani2017attention} to capture the longer-term audio context and sufficiently learn correlated audio-emotion features for expression coefficient prediction, which involves precise facial emotion and synchronized lip movement. Notably, a learned intensity token is extended to control the emotion intensity continuously. Furthermore, to generate high-resolution realistic faces, we propose a coarse-to-fine style-based identity-generalized model, High-fidelity Emotional Face (HEF) generator, which integrates appearance features, geometry information, and a style code within an elegant design. As shown in Fig.~\ref{fig:teaser}(a), unlike the EAMM that predicts the flow field at a single resolution by an isolated process, we hierarchically perform the flow estimation in a residual manner and incorporate it with texture refinement for efficiency.

In summary, we make the following three contributions:

\begin{itemize}

\item We propose a novel AME that provides a unified multi-modal semantic-rich emotion space, allowing flexible emotion control and unseen emotion generalization, which is the first attempt in this field.

\item We propose a novel HEF to hierarchically learn the facial deformation by sufficiently modeling the interaction among emotion, source appearance, and drive geometry for the high-resolution one-shot generation.

\item Abundant experiments are conducted to demonstrate the superiority of our method for flexible and generalized emotion control, and high-resolution one-shot talking face animation over SOTA methods.
% generating emotional audio-driven talking faces over SOTA methods.

\end{itemize}

\begin{figure*}[t!]
	\centering
	\includegraphics[width=0.95\textwidth]{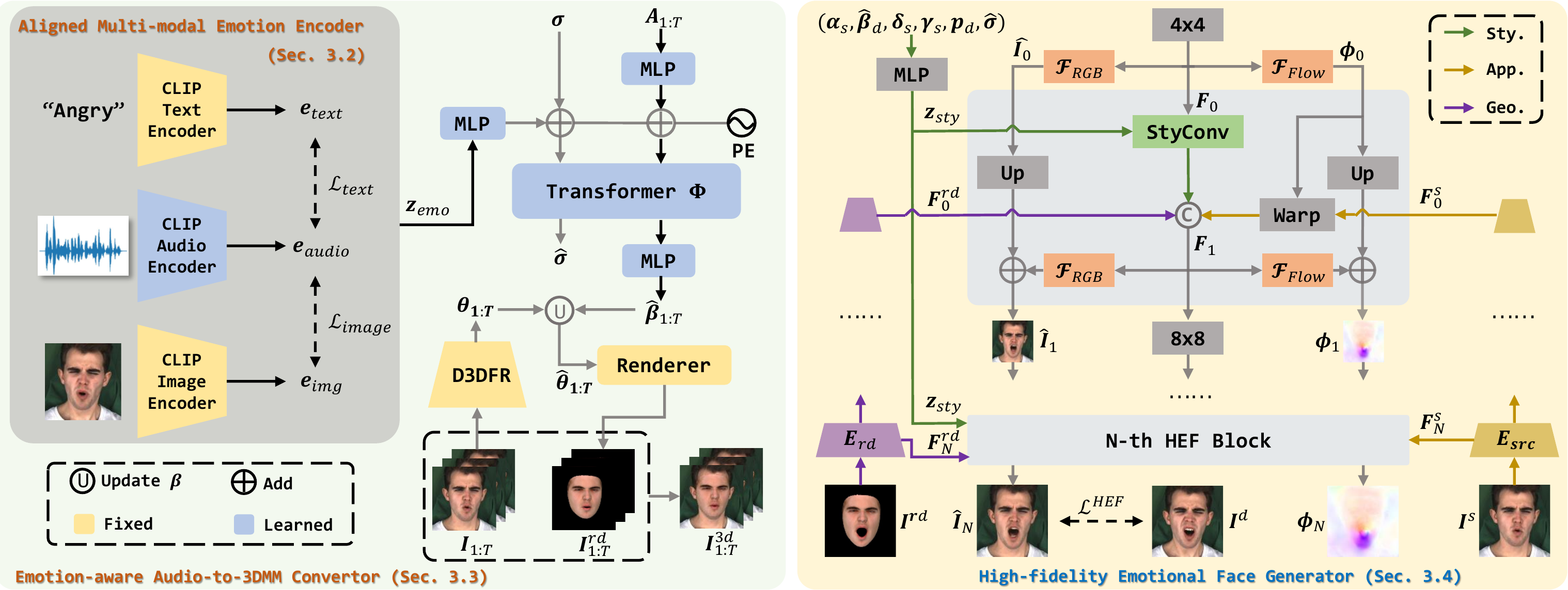}
	\caption{\textbf{Overview of the proposed method.} Our method is a two-stage framework that transfers the audio content and multi-modal emotion sources to a static portrait $\boldsymbol{I}^{s}$. Specifically,
	the Emotion-aware Audio-to-3DMM Convertor encodes MFCC sequence $\boldsymbol{A}_{1:T}$, the emotion style $\boldsymbol{z}_{emo}$ embedded from Aligned Multi-modal Emotion encoder, and a learnable intensity token $\boldsymbol{\sigma}$ to predict the expression coefficient sequence $\hat{\boldsymbol{\beta}}_{1:T}$. The followed High-fidelity Emotional Face Generator receives the style vectors $\boldsymbol{z}_{sty}$ (in \textcolor{green}{Green}) mapped from the modified coefficients and updated intensity token $\hat{\boldsymbol{\sigma}}$, the source appearance $\boldsymbol{F}^{s}_{0:N}$ (in \textcolor{olive}{Olive}) from $\boldsymbol{I}^{s}$, and the driving geometry $\boldsymbol{F}^{rd}_{0:N}$ (in \textcolor{violet}{Violet}) from $\boldsymbol{I}^{rd}$, to hierarchically generate the facial deformation $\boldsymbol{\phi}_{0:N}$ to guide the animated emotional talking face synthesis.
	}
	\label{fig:pipeline}
\end{figure*}

\section{Related Work}

\subsection{Audio-driven Talking Face Generation} Early efforts focus on modeling the audio information into pure latent feature space~\cite{chung2017you, zhou2019talking, zhou2021pose, zhang2020apb2facev2} and employing a conditioned image generation framework~\cite{goodfellow2014generative, mirza2014conditional} to synthesize realistic faces. Recently, structural information has been leveraged as the explicit intermediate representation to bridge the gap between audio and visual domains. Das \emph{et al.}~\cite{das2020speech} capture facial motion in landmarks\cite{zhang2020apb2face} from the given audio sequence and then synthesizes texture conditioned on these structures. Considering the attributes are entangled within 2D landmarks, 3DMM~\cite{deng2019accurate} is introduced in this task~\cite{ren2021pirenderer, wu2021imitating, yi2020audio, zhang2021facial, ye2022dynamic, zhang2021flow, yin2022styleheat}. Specifically, FACIAL~\cite{zhang2021facial} predicts head pose, expression, and AU45 from deep speech features by the fully connected networks. PIRenderer~\cite{ren2021pirenderer} adopts LSTMs~\cite{hochreiter1997long} to autoregressively deduce expressions and poses while LSP~\cite{lu2021live} employs GRUs~\cite{cho2014learning}. We follow these methods but use the non-autoregressive Transformer~\cite{vaswani2017attention} to capture the long-term audio context and provide the sequence-level representations for more accurate coefficients regression, which helps exhibit precise emotion in the texture generator.

\subsection{Emotion Conditioned Generation} Early efforts~\cite{pumarola2018ganimation, ding2018exprgan} serve this task as the domain transfer, but they fail to synchronize lip movement with speech. Recently, MEAD~\cite{wang2020mead} releases a high-quality talking head video dataset with annotations of emotion category and intensity. Subsequent works~\cite{wang2020mead, ye2022dynamic} encode the expression labels in one-hot vectors to maintain the desired expression. EVP~\cite{ji2021audio} decomposes audio into the corresponding emotion style to capture more semantic information. However, these works are in identity-specific design. Consequently, recent GC-AVT~\cite{liang2022expressive} and EAMM~\cite{ji2022eamm} explore one-shot setting and drive facial expressions by reference faces, but these methods only support emotion style obtained from a single modality and struggle to produce high-resolution faces. In contrast, we construct multi-modal emotion sources into a unified feature space, supporting \textit{diverse} modalities within a single model. Besides, our hierarchical texture generator could produce \textit{high-resolution} faces with the desired appearance, pose, and expression.

\subsection{CLIP-guided Synthesis} CLIP~\cite{radford2021learning} is perfect for visual tasks with textual assistance, which has proven effective for image editing~\cite{frans2021clipdraw, schaldenbrand2021styleclipdraw, patashnik2021styleclip, wei2022hairclip, kim2021diffusionclip}, domain transfer~\cite{kwon2021clipstyler, liu2022name}, and 3D avatar~\cite{tevet2022motionclip, hong2022avatarclip}. Besides, some works~\cite{baltruvsaitis2018multimodal, zhao2021emotion} have verified the necessity of employing multi-modal information. In this work, we supplement the emotion style in text prompt and unify the multi-modal features in the CLIP space, which contains rich semantics and unprecedented textual and visual understanding ability.
Once trained, our method could generalize to \textit{unseen} emotion styles located in similar emotion domains of CLIP, which is not considered in the previous emotional talking face generation methods.

\section{Method}
The emotional talking face generation aims at driving the source face by the given emotion style and audio content. The desired framework for this task should embody several core properties: 1) Since several modalities could represent emotion style, the designed method should support diverse modalities within one single model to achieve a flexible and user-friendly interaction. 2) The trained network could be applied for unseen emotion styles and identities, and generates high-resolution realistic faces.
To achieve the above goals, we first design an Aligned Multi-modal Emotion (AME) encoder to produce a unified feature space in Sec.~\ref{sec:3.2}. Then a Transformer-based Emotion-aware Audio-to-3DMM Convertor (EAC) receives emotion style from AME, along with given audio, to connect audio-emotion inputs with the 3DMM (Sec.~\ref{sec:3.3}). Finally, we propose a style-based High-fidelity Emotional Face (HEF) generator to synthesize the realistic emotional talking faces of arbitrary identities by learning hierarchical facial deformation (Sec.~\ref{sec:3.4}). Our pipeline is depicted in Fig.~\ref{fig:pipeline}. 
% We will explain each component in detail as follows.

% Specifically, EAC takes a given Mel-frequency Cepstral Coefficients (MFCC) clips $\boldsymbol{A}=\left\{\boldsymbol{A}_1, \ldots, \boldsymbol{A}_T\right\}$, a learnable emotion intensity token $\boldsymbol{\sigma}$, and the emotion style $\boldsymbol{z}_{emo}$ output from AME as input, to fully model the correlated audio-emotion features for accurate expression coefficients prediction, obtaining $\hat{\boldsymbol{\beta}}=\left\{\hat{\boldsymbol{\beta}_1}, \ldots, \hat{\boldsymbol{\beta}_T}\right\}$ and an updated intensity token $\hat{\boldsymbol{\sigma}}$. In the second stage, the updated intensity token $\hat{\boldsymbol{\sigma}}$ and the recombined 3DMM are embedded to the latent style code $\boldsymbol{z}_{sty}$, which contains rich emotion and geometrical information. The encoder $\boldsymbol{E}_{src}$ and $\boldsymbol{E}_{rd}$ extract appearance features $\boldsymbol{F}^{s} = \left\{\boldsymbol{F}_1^{s}, \ldots, \boldsymbol{F}_T^{s}\right\}$ and $\boldsymbol{F}^{rd} = \left\{\boldsymbol{F}_1^{rd}, \ldots, \boldsymbol{F}_T^{rd}\right\}$ simultaneously. The above three elements are input to the generator $\boldsymbol{G}$ to synthesize the realistic faces. In the following, we proceed to discuss the two stages in detail.

\subsection{3D Face Descriptors}
\label{sec:3.1}
Following the previous works, we employ 3DMM parameters as the intermediate representation. With 3DMM, the 3D shape $\mathbf{S}$ and albedo texture $\mathbf{T}$ are parameterized as:

\begin{equation}
  \begin{aligned}
    \mathbf{S} &= \bar{\mathbf{S}} + \mathbf{B}_{id} \boldsymbol{\alpha} + \mathbf{B}_{exp} \boldsymbol{\beta}, \\
    \mathbf{T} &= \bar{\mathbf{T}} + \mathbf{B}_{t} \boldsymbol{\delta},
  \end{aligned}
\end{equation}
where $\bar{\mathbf{S}}$ and $\bar{\mathbf{T}}$ denote the mean face shape and albedo texture. $\mathbf{B}_{id}$, $\mathbf{B}_{exp}$, and $\mathbf{B}_{t}$ are the bases of identity, expression, and the texture computed via Principal Component Analysis (PCA). Coefficients $\boldsymbol{\theta} = \left\{\boldsymbol{\alpha} \in \mathbb{R}^{80}, \boldsymbol{\beta} \in \mathbb{R}^{64}, \boldsymbol{\delta} \in \mathbb{R}^{80}, \boldsymbol{\gamma} \in \mathbb{R}^{27}, \boldsymbol{p} \in \mathbb{R}^{6}\right\}$ describe the identity, expression, texture, illumination, pose, respectively. Although off-the-shelf 3D face reconstruction model D3DFR~\cite{deng2019accurate} could capture relatively accurate facial features, they fail to produce reliable expression coefficients for extreme emotional faces due to the lack of tailored training on the corresponding dataset. Consequently, we do not directly adopt the extracted expression coefficients $\boldsymbol{\beta}$ as the constraint for the EAC training (Sec.~\ref{sec:3.3} $\mathcal{L}_{emo}$).

\subsection{Aligned Multi-modal Emotion Encoder}
\label{sec:3.2}
To unify the emotion conditions from the text, audio, and image domains within one framework, we naturally choose CLIP space as the multi-domain feature space. Specifically, we design an \textit{Aligned Multi-modal Emotion Encoder}, which consists of the fixed CLIP text and image encoders, and the learned CLIP audio encoder to produce emotion embedding $\boldsymbol{e}_{text}$, $\boldsymbol{e}_{img}$, and $\boldsymbol{e}_{audio}$. In practice, the CLIP audio encoder is the basic Transformer-based architecture with a \texttt{Cls} token for pool purposes and learns to embed $\boldsymbol{e}_{audio}$. AME receives the synchronized multi-modal inputs during training, and the output emotion code $\boldsymbol{z}_{emo}$ is the combination of the above three embedding along the batch dimension, in which each modality shares the \textit{same} ground truth. Thus it allows pixel-level constraints on the generated face. To facilitate unified feature space learning and emotion disentanglement from the entangled audio, we further apply a feature-level loss to align the $\boldsymbol{e}_{audio}$ to CLIP textual and visual space simultaneously. 
For testing, we could adopt arbitrary emotion embedding as $\boldsymbol{z}_{emo}$, which is more flexible in applications:
\begin{equation}
    \begin{aligned}
      \boldsymbol{z}_{emo} &= \left[\boldsymbol{e}_{text},\boldsymbol{e}_{audio},\boldsymbol{e}_{img}\right],\text{at training stage}\\
      \boldsymbol{z}_{emo} &\in \left\{\boldsymbol{e}_{text},\boldsymbol{e}_{audio},\boldsymbol{e}_{img}\right\},\text{at test stage}
    \end{aligned}
\end{equation}
where $[\cdot]$ means concatenation. The merits of aligning multi-modal emotion features to CLIP space are two-folder: First, our model distills the emotion cues from the CLIP domain and inherits rich semantic knowledge to benefit unseen emotion generalization. Second, CLIP already provides shared textural and visual feature space, which is easier to train a single audio encoder than the whole network.

\subsection{Emotion-aware Audio-to-3DMM Convertor}
\label{sec:3.3}

\noindent\textbf{Architecture.} To project the audio content and emotion style to expression coefficients of 3DMM, we propose a Transformer-based \textit{Emotion-aware Audio-to-3DMM Convertor}. As shown in Fig.~\ref{fig:pipeline}, $\boldsymbol{A}$ provides the information of lip movement, and $\boldsymbol{z}_{emo}$ is the emotion embedding. Besides, instead of utilizing a one-hot coding to control the emotion intensity~\cite{wang2020mead}, we prepend a learnable intensity token $\boldsymbol{\sigma}$ inspired by the ViT~\cite{dosovitskiy2020image}. This token is the product of the base learnable intensity vector and the intensity scalar: $\boldsymbol{\sigma} = \mu \boldsymbol{\sigma}_{base}$,
% \begin{equation}
%   \begin{aligned}
%     \boldsymbol{\sigma} &= \mu \boldsymbol{\sigma}_{base},
%   \end{aligned}
% \end{equation}
where $\mu \in \left\{1,2,3\right\}$ during training, corresponds to the ground-truth intensity annotated in MEAD. It can be a random value range from 1 to 3 during testing. In practice, we map the audio feature dimension and concatenate them with intensity token $\boldsymbol{\sigma}$. The MLP is used to initially separate emotion cues from CLIP space. The above are added with positional embedding $\text{PE}$ and fed into the Transformer $\boldsymbol{\Phi}$ for expression coefficients prediction:
\begin{eqnarray}
%   \begin{aligned}
    &\hat{\boldsymbol{\sigma}}, \bar{\boldsymbol{\beta}} = \boldsymbol{\Phi}([\boldsymbol{\sigma},\text{MLP}(\boldsymbol{A})] + \text{PE} + \text{MLP}(\boldsymbol{z}_{emo})), \\
    &\hat{\boldsymbol{\beta}} = \text{MLP}(\bar{\boldsymbol{\beta}}).
%   \end{aligned}
\end{eqnarray}

\noindent\textbf{Objectives.} We train this stage by using five losses:
\begin{equation}
  \begin{aligned}
    \mathcal{L}^{EAC} &=
     \lambda_{clip}^{EAC}\mathcal{L}_{clip} + \lambda_{emo}^{EAC}\mathcal{L}_{emo} + \lambda_{rec}^{EAC}\mathcal{L}_{rec}\\ &+ \lambda_{lm}^{EAC}\mathcal{L}_{lm}
    + \lambda_{reg}^{EAC}\mathcal{L}_{reg}.
  \end{aligned}
\end{equation}

\noindent\textit{Clip Loss} $\mathcal{L}_{clip}$: As stated in Sec.~\ref{sec:3.2}, we force the emotion feature from the audio close to that from the text and image by using cosine distance: $\mathcal{L}_{image} = 1-\text{cos}(\boldsymbol{e}_{img}, \boldsymbol{e}_{audio})$, $\mathcal{L}_{text} = 1-\text{cos}(\boldsymbol{e}_{text}, \boldsymbol{e}_{audio})$, $\mathcal{L}_{clip} = \mathcal{L}_{image} + \mathcal{L}_{text}$.
% \begin{equation}
%     \begin{aligned}
%         \mathcal{L}_{clip} = \mathcal{L}_{image} + \mathcal{L}_{text}.
%     \end{aligned}
% \end{equation}
% Different from GC-AVT applying contrastive loss for audio-visual content synchronization, we directly adopt cosine distance and focus on multi-modal emotion style alignment.

\noindent\textit{Emotion Consistency Loss} $\mathcal{L}_{emo}$: This is a critical term to distill and infuse the semantic emotion representation of CLIP. Due to the unreliable of extracted expression coefficients (Sec.~\ref{sec:3.1}), we turn to the image level for help, projecting the modified 3DMM onto the 2D image plane with a differentiable renderer: $\boldsymbol{R}(\hat{\boldsymbol{\theta}}) \rightarrow \boldsymbol{I}^{rd}$, which is then blended to the original face by the face mask $\boldsymbol{M}$ output from $\boldsymbol{R}$: $\boldsymbol{I}^{3d} = \boldsymbol{M} \odot \boldsymbol{I}^{rd} + (1-\boldsymbol{M}) \odot \boldsymbol{I}.$
% \begin{equation}
%   \begin{aligned}
%     \boldsymbol{I}^{3d} &= \boldsymbol{M} \odot \boldsymbol{I}^{rd} + (1-\boldsymbol{M}) \odot \boldsymbol{I}.
%   \end{aligned}
% \end{equation}
After that, we adopt an emotion recognition network~\cite{mollahosseini2017affectnet} to compute the perceptual difference between the input and rendered images:
\begin{equation}
  \begin{aligned}
    \mathcal{L}_{emo}= \left\|\boldsymbol{\varphi}_{emo}(\boldsymbol{I}^{3d})-\boldsymbol{\varphi}_{emo}(\boldsymbol{I})\right\|_{2},
  \end{aligned}\label{eq:emo}
\end{equation}
where $\boldsymbol{\varphi}_{emo}$ is the backbone before the last linear layer.

\noindent\textit{Reconstruction Loss} $\mathcal{L}_{rec}$: We compute the pixel level loss
between the input and the rendered images on the face area: $\mathcal{L}_{rec}= \left\|\boldsymbol{I}^{rd}-\boldsymbol{M} \odot \boldsymbol{I}\right\|_{2}.$
% \begin{equation}
%   \begin{aligned}
%     \mathcal{L}_{rec}= \left\|\boldsymbol{I}^{rd}-\boldsymbol{M} \odot \boldsymbol{I}\right\|_{2}.
%   \end{aligned}
% \end{equation}

\noindent\textit{Landmark Loss} $\mathcal{L}_{lm}$: We predict 68 points from the original 3DMM $\boldsymbol{\theta}$ and the modified one $\hat{\boldsymbol{\theta}}$, obtaining $\boldsymbol{l}$ and $\hat{\boldsymbol{l}}$. $\mathcal{L}_2$ distance is used to measure them: $\mathcal{L}_{lm}= \left\|\boldsymbol{l}-\hat{\boldsymbol{l}}\right\|_{2}.$
% \begin{equation}
%   \begin{aligned}
%     \mathcal{L}_{lm}= \left\|\boldsymbol{l}-\hat{\boldsymbol{l}}\right\|_{2}.
%   \end{aligned}
% \end{equation}

\noindent\textit{Expression Regularization Loss} $\mathcal{L}_{reg}$: This term is used to smooth the training phase, calculating the distance between the $\boldsymbol{\beta}$ and $\hat{\boldsymbol{\beta}}$ with a small weight: $\mathcal{L}_{reg}= \left\|\boldsymbol{\beta}-\hat{\boldsymbol{\beta}}\right\|_{2}.$
% \begin{equation}
%   \begin{aligned}
%     \mathcal{L}_{reg}= \left\|\boldsymbol{\beta}-\hat{\boldsymbol{\beta}}\right\|_{2}.
%   \end{aligned}
% \end{equation}

\subsection{High-fidelity Emotional Face Generation}
\label{sec:3.4}

\noindent\textbf{Architecture.} 
As shown in Fig.~\ref{fig:teaser}(a), GC-AVT does not generate consistent texture and background with the source, while EAMM relies on aligned inputs and produces blurred results. Both are in low-resolution and poor quality. Thus, we carefully modify StyleGAN2~\cite{karras2020analyzing} and propose \textit{High-fidelity Emotional Face Generation}. Specifically, as shown in Fig.~\ref{fig:pipeline}, we randomly sample a driving face $\boldsymbol{I}^d$ from given clips and a face with the same identity but different emotion as the source $\boldsymbol{I}^s$. To transfer the audio-synchronized lip movement, pose, and expression from the drive to the source face, the style code is defined as:
\begin{equation}
  \begin{aligned}
     \boldsymbol{z}_{sty}=\text{Linear}([\boldsymbol{\alpha}_s, \hat{\boldsymbol{\beta}_d}, \boldsymbol{\delta}_s, \boldsymbol{\gamma}_s, \boldsymbol{p}_d, \hat{\boldsymbol{\sigma}}]).
  \end{aligned}
\end{equation}

In addition to $\boldsymbol{z}_{sty}$, $\boldsymbol{E}_{src}$ extracts pyramid appearance features $\boldsymbol{F}^{s}_{0:N}$ from the $\boldsymbol{I}^s$, where $N$ is the number of HEF blocks. $\boldsymbol{E}_{rd}$ simultaneously provides hierarchical features $\boldsymbol{F}^{rd}_{0:N}$, which embody the emotional textures and geometrical guidance of the desired face from the $\boldsymbol{I}^{rd}$. Thus we have three faithful implicit-explicit inputs for HEF. Furthermore, to align the $\boldsymbol{F}^{s}_{i}$ and $\boldsymbol{F}^{rd}_{i}$, different from existing methods~\cite{ren2021pirenderer, siarohin2019first, ji2022eamm, xu2022designing} employing isolated flow fields prediction module, we simultaneously estimate the spatial deformations and refine the final images in a residual manner within each HEF block. As depicted in Fig.~\ref{fig:pipeline}, we have:
\begin{eqnarray}
    &\boldsymbol{F}_{i+1} = [\boldsymbol{\mathcal{T}}(\boldsymbol{\phi}_i, \boldsymbol{F}_{i}^s), \mathbf{StyleConv}(\boldsymbol{F}_{i}, \boldsymbol{z}_{sty}), \boldsymbol{F}^{rd}_{i}], \\
    &\boldsymbol{\phi}_{i+1} = \boldsymbol{\mathcal{F}}_{flow}(\boldsymbol{F}_{i+1}) + \mathbf{Up}(\boldsymbol{\phi}_{i}), \\
    &\hat{\boldsymbol{I}}_{i+1} = \boldsymbol{\mathcal{F}}_{rgb}(\boldsymbol{F}_{i+1}) +  \mathbf{Up}(\hat{\boldsymbol{I}}_{i}),
\end{eqnarray}
where $\mathbf{StyleConv}$ denotes the style convolution in StyleGAN2 ($\mathbf{Up}$ + Conv$3\times3$ with modulation). Please refer to its paper for more details. $\mathbf{Up}$ means upsampling, $\boldsymbol{\mathcal{T}}$ denotes warping operation, $\boldsymbol{\mathcal{F}}_{flow}$ converts the high-dimensional features to dense flow fields, and $\boldsymbol{\mathcal{F}}_{rgb}$ to realistic RGB images, respectively.

\noindent\textbf{Objectives.} We employ three loss terms to measure the difference between $\boldsymbol{I}^d$ and $\hat{\boldsymbol{I}}_{N}$ at the pixel and perceptual level by a \textit{Reconstruction Loss} $\mathcal{L}_{rec}$ as $\mathcal{L}_1$ distance and a \textit{Perceptual Loss} $\mathcal{L}_p$ as the LPIPS loss~\cite{zhang2018unreasonable}. Besides, we adopt \textit{Adversarial Loss} to ensure the authenticity of the generated faces. The overall objective is a combination of the above:
% We employ four loss terms for training HEF, \emph{i.e.}, a emotion consistency loss $\mathcal{L}_{emo}$ measure the emotion perceptual difference between $\hat{\boldsymbol{I}}$ and $\hat{\boldsymbol{I}}_{d}$ as Eq.\ref{eq:emo}, a reconstruction loss $\mathcal{L}_{rec}$ , a perceptual loss $\mathcal{L}_p$, and a adversarial loss $\mathcal{L}_{adv}$. The total loss is defined as follow:
\begin{equation}
  \begin{aligned}
    \mathcal{L}^{HEF} = \lambda_{rec}^{HEF}\mathcal{L}_{rec} + \lambda_{p}^{HEF}\mathcal{L}_{p} + \lambda_{adv}^{HEF}\mathcal{L}_{adv}.
  \end{aligned}
\end{equation}

\section{Experiments}
\subsection{Datasets and Implementation Details}

\noindent\textbf{Datasets.} Our model is trained on MEAD~\cite{wang2020mead} with eight expression types (neutral, angry, contempt, disgusted, fear, happy, sad, and surprised) and three intensity levels (levels 1, 2, 3), which contains fine-grained emotion annotation, helping distill emotion space from CLIP for unseen style generalization. We randomly select 36 identities of front-view videos for training and the rest identities for testing.
% We adopt another dataset RAVDESS~\cite{livingstone2018ryerson} for further comparison. Please refer to supplement materials for more details.

\noindent\textbf{Metrics.} We adopt PSNR, SSIM~\cite{wang2004image}, and FID~\cite{heusel2017gans} to evaluate the quality of generated images. We use Landmarks Distance (LMD)~\cite{chen2018lip} around the mouth and the confidence score (Sync) proposed in SyncNet~\cite{chung2016out} to measure the accuracy of mouth shapes and lip synchronization. We compute Cosine Similarity (CSIM) to evaluate identity preservation. CurricularFace~\cite{huang2020curricularface} is used to extract identity embedding. We use Emotion Feature Distance (EFD) to measure the accuracy of the emotion representation, which is extracted by FAN~\cite{meng2019frame} different from the model in the loss calculation.

\noindent\textbf{Implementation Details.} The EAC and HEF are trained independently. For EAC, we randomly sample consecutive $T=32$ clips for training. The values of the loss weights are set to $\lambda_{clip}^{EAC}=1$, $\lambda_{emo}^{EAC}=1$, $\lambda_{rec}^{EAC}=100$, $\lambda_{lm}^{EAC}=0.1$, $\lambda_{reg}^{EAC}=0.01$. We use a learning rate of 0.0002 and 128 batch size to train EAC with the Adam optimizer on one V100 GPU. For HEF, we fix EAC and the values of the loss weights are set to $\lambda_{rec}^{HEF}=5$, $\lambda_{p}^{HEF}=5$, $\lambda_{adv}^{HEF}=1$. This training phase also adopts Adam optimizer with 0.002 learning rate, using 8 V100 GPUs and 1 clip (8 images) per GPU. The output image size of HEF is $512 \times 512$ with $N=7$ blocks. The audios are pre-processed to 16kHz, then converted to mel-spectrograms with FFT window size 1280, hop length 160, and 80 Mel filter-bank as PC-AVS~\cite{zhou2021pose}.

\begin{figure}
	\centering
	\includegraphics[width=0.45\textwidth]{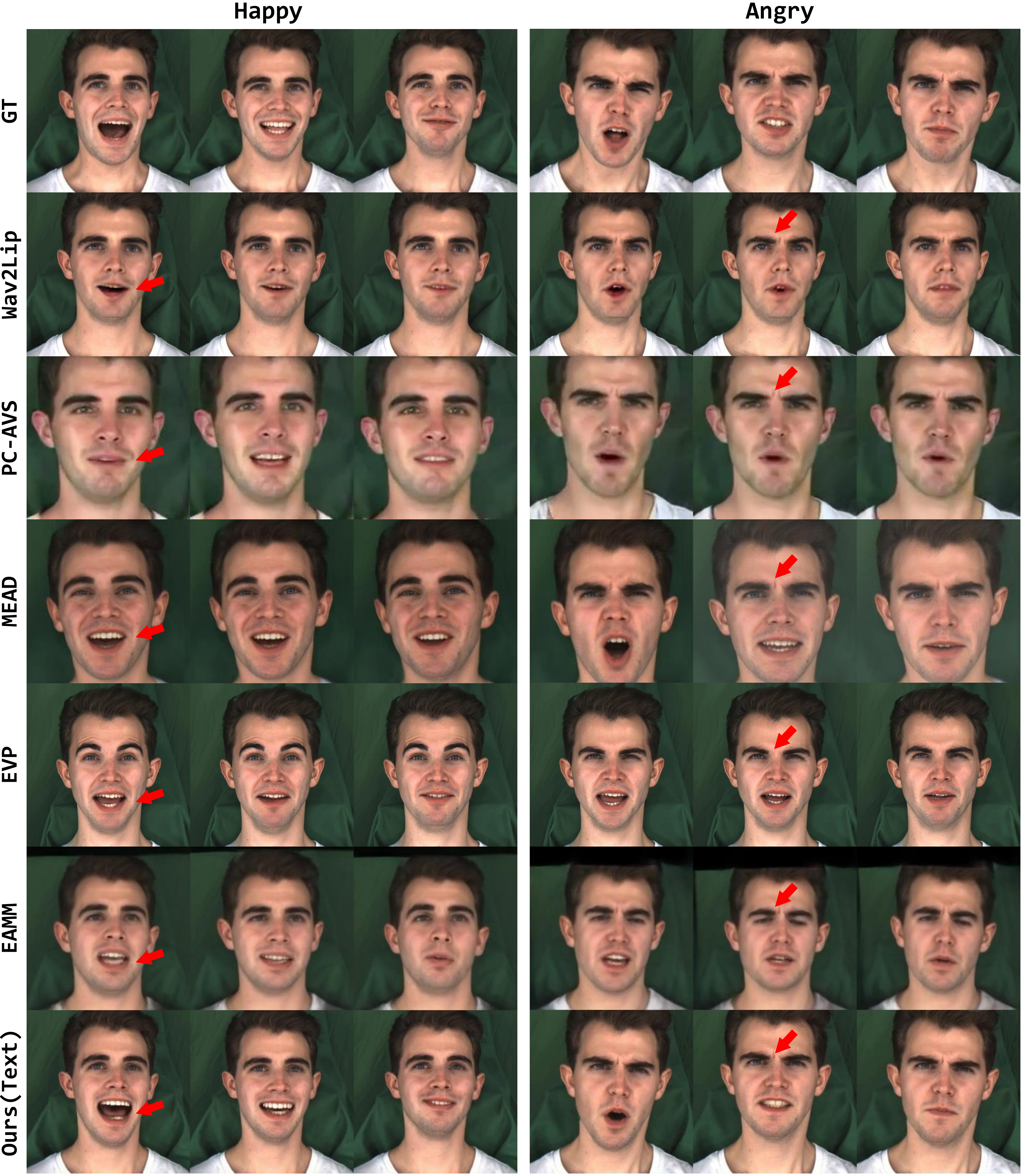}
	\caption{Qualitative results on MEAD dataset. Different columns mean several sampled timestamps (\textit{same as the following figures}). Images are from officially released codes for fair comparisons.
	}
	\label{fig:sota}
\end{figure}

\begin{figure}[t!]
	\centering
	\includegraphics[width=0.45\textwidth]{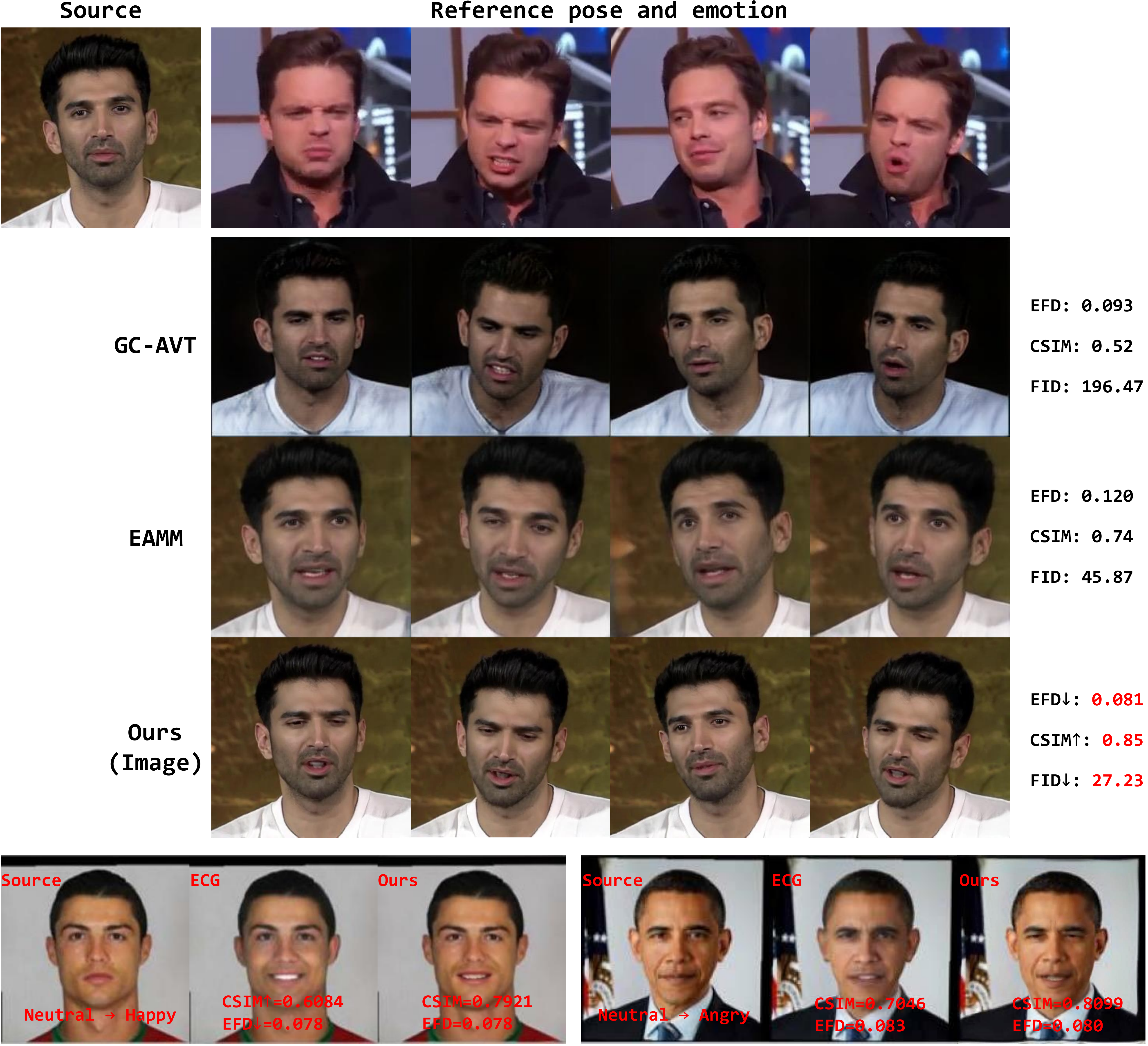}
	\caption{Qualitative comparison with GC-AVT, EAMM, and ECG. The top part is sampled from Fig. 3 of GC-AVT. The bottom part is sampled from Fig. 4 of ECG. Quantitative results of these cases are attached in the figure. We ignore the metric of mouth shape because the audios for these sequences are not available.}
	\label{fig:sota2}
\end{figure}

\subsection{Comparison with State-of-the-Arts}
\label{sec:4.2}

\noindent\textbf{Qualitative Results.} We show the results of M003 to ensure that each method could animate the source face, and this identity is not in our training set. The first frame of each test video as the source and its audio, label, and a random face from the same video for multi-modal emotion conditions. As shown in Fig.~\ref{fig:sota}, we select three frames of two emotion styles for comparison. It can be seen that common audio-driven methods, Wav2Lip~\cite{prajwal2020lip} and PC-AVS, struggle to generate desired emotions with synchronized lip shapes, while the synthesized images from MEAD are of poor quality. EVP and EAMM suffer identity inconsistency with the source and show less rich expression due to lacking intensity modeling. In contrast, benefiting from sufficient emotion semantics and intensity learning, our method with text as the emotion condition produces more accurate expressions. Besides, our results show more realistic textures than all competitors due to coarse-to-fine flow field and image refining. We further compare our method with GC-AVT, EAMM, and ECG~\cite{sinha2022emotion}. As shown in Fig.~\ref{fig:sota2}, since GC-AVT does not release codes, we adopt the officially attached results in its paper and employ image as our emotion condition for a fair comparison. In terms of emotion accuracy, identity consistency, and image quality, our method obviously outperform these SOTA methods. We also attach the corresponding quantitative results of this case on EFD, CSIM, and FID in Fig.~\ref{fig:sota2}, which are consistent with the qualitative results. The same conclusion could deduced from the bottom part of Fig.~\ref{fig:sota2} when compared with ECG.

\begin{table}[t]
   \centering
   \scriptsize
   \renewcommand\arraystretch{1.0}
   \setlength\tabcolsep{4pt}
   \begin{tabular}{C{25pt}C{19pt}C{22pt}C{20pt}C{23pt}C{19pt}C{24pt}C{23pt}}
      \toprule
      Method & EFD $\downarrow$ & LMD $\downarrow$ & Sync $\uparrow$ & CSIM $\uparrow$ & FID $\downarrow$ & PSNR $\uparrow$ & SSIM $\uparrow$ \\
      \midrule
      Wav2Lip & 0.112  & 2.59   & 3.26   & 0.82	            & 20.15              & 29.22             & 0.70              \\
      PC-AVS & 0.110   & 2.68   & 3.12   & 0.80	            & 29.55              & 28.97             & 0.68              \\
      \midrule
      MEAD & 0.084  & 2.62   & 3.09   & 0.81	            & 30.69              & 28.48             & 0.65              \\
      EVP & 0.106   & 2.54   & 3.21   & 0.70	            & \textbf{12.83}              & 29.67             & 0.73              \\
      EAMM & 0.092   & 2.50   & 3.26   & 0.74	            & 29.01              & 29.33             & 0.75              \\
      \midrule
      Ours-A & \underline{0.069}  & 2.36    & 3.50   & \underline{0.83}	            & 15.91              & \underline{30.09}             & 0.85  \\
      Ours-I & 0.071  & 2.36   & 3.53    & \textbf{0.84}	& \underline{15.89}  & \textbf{30.10} & \underline{0.87}              \\
      Ours-T & \textbf{0.065}  & \textbf{2.31}    & \textbf{3.57}    & \textbf{0.84}	   & 15.90     & \underline{30.09}    & \textbf{0.88}     \\
    %   Ours-All & \textbf{0.065}  & \underline{2.33}    & \underline{3.54}    & \textbf{0.84}	   & \textbf{15.87}     & \underline{30.09}    & \textbf{0.88}     \\
      \bottomrule
   \end{tabular}
   \caption{Quantitative comparison on MEAD dataset. Ours-A, -I, and -T mean audio, image, and text, respectively.}
   \label{tab:sota}
\end{table}

\noindent\textbf{Quantitative Results.} We adopt several metrics to evaluate the superiority of our approach on image quality, landmark accuracy, lip synchronization, identity preservation, and emotion accuracy. As shown in Tab.~\ref{tab:sota}, our method outperforms most metrics except for the FID. EVP achieves higher FID but exhibits a weak manipulated ability, which can be inferred from the lower CSIM and Sync, and higher LMD. Besides, comparing the last three rows, we observe that emotion modality mainly affects structural metrics, and text-driven results show better performance than that of audio- and image-driven in most metrics, which may attribute to the better emotion disentanglement of text prompts. Thus \textit{we use the text as our default emotion condition} in the following experiments. We further report the user study in supplementary materials, specially evaluating the overall quality, the generalization to unseen emotions, and the video temporal consistency.

\subsection{Further Analysis}

\begin{figure}[t]
	\centering
	\includegraphics[width=0.4\textwidth]{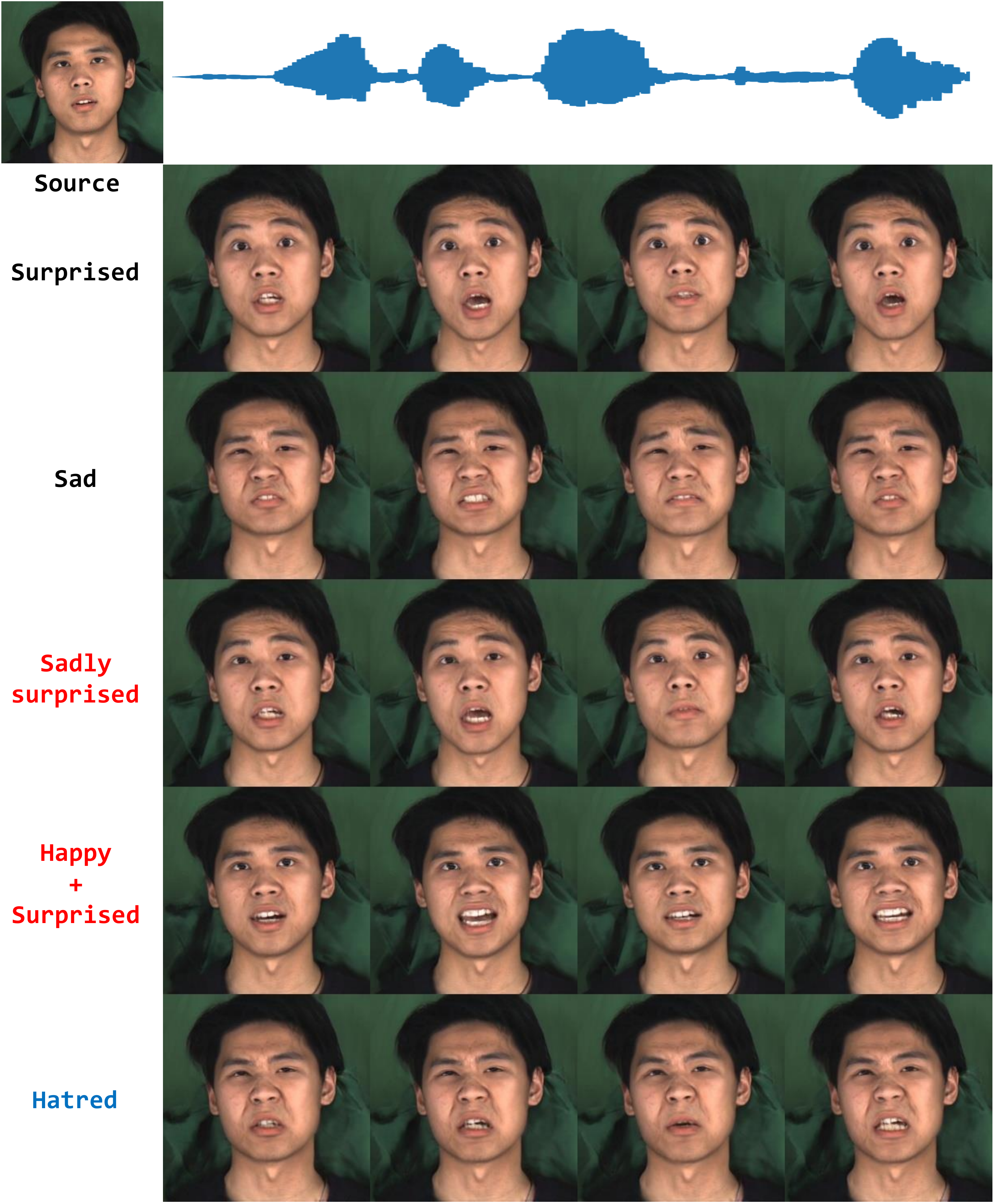}
	\caption{Results of unseen emotion styles. Rows 4 and 5 (in \textcolor{red}{Red}) are the compound styles, and row 6 (in \textcolor{blue}{Blue}) is a totally new style. 
	}
	\label{fig:unseen_emo}
\end{figure}

\begin{figure}[t!]
	\centering
	\includegraphics[width=0.42\textwidth]{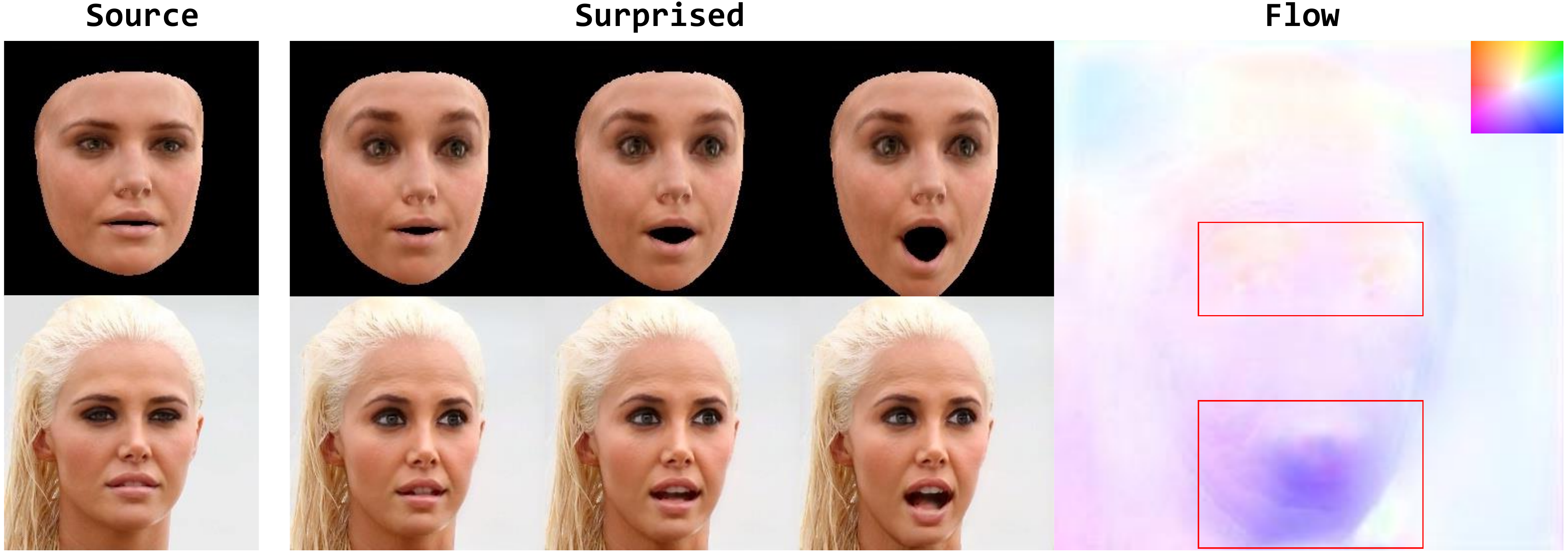}
	\caption{Results of unseen identity. We visualize the rendered images, final outputs, and predicted flow fields. The color wheel of flow fields is attached on the top right for reference.
	}
	\label{fig:unseen_id}
\end{figure}

\noindent\textbf{Generalizing to Unseen Emotion Styles.} Unseen emotion styles include compound and totally new styles. As shown in Fig.~\ref{fig:unseen_emo}, rows 2 and 3 are the basic styles, row 4 shows the results of the given \textit{Sadly surprised}, and row 5 of the average embedding of \textit{Happy} and \textit{Surprised}, which indicates the flexible manipulation for compound emotion. We further present the new style \textit{Hatred} in the sixth row. The correct exhibition of these unseen styles verifies the flexibility and rich semantic priors of the CLIP feature space.

\noindent\textbf{Generalizing to Unseen Identities.} As shown in Fig.~\ref{fig:teaser} and Fig.~\ref{fig:sota2}, our method trained on MEAD could generalize to unseen identities from VoxCeleb2~\cite{chung2018voxceleb2}. We further conduct a qualitative visualization in Fig.~\ref{fig:unseen_id}. Specifically, we sample a face from CelebA-HQ as an unseen identity. The generated faces preserve the identity and exhibit desired emotion, reflecting both on realistic and rendered faces. Besides, we attach the flow map predicted from HEF in the fifth column. Please pay attention to facial movements, especially in the mouth and eyes. We can conclude that HEF accurately models the emotion-related facial movement conditioned on the intermediate structure, which is not sensitive to the identity textures. Thus MEAD is sufficient to provide diverse movements for training.

\begin{figure}[t!]
	\centering
	\includegraphics[width=0.45\textwidth]{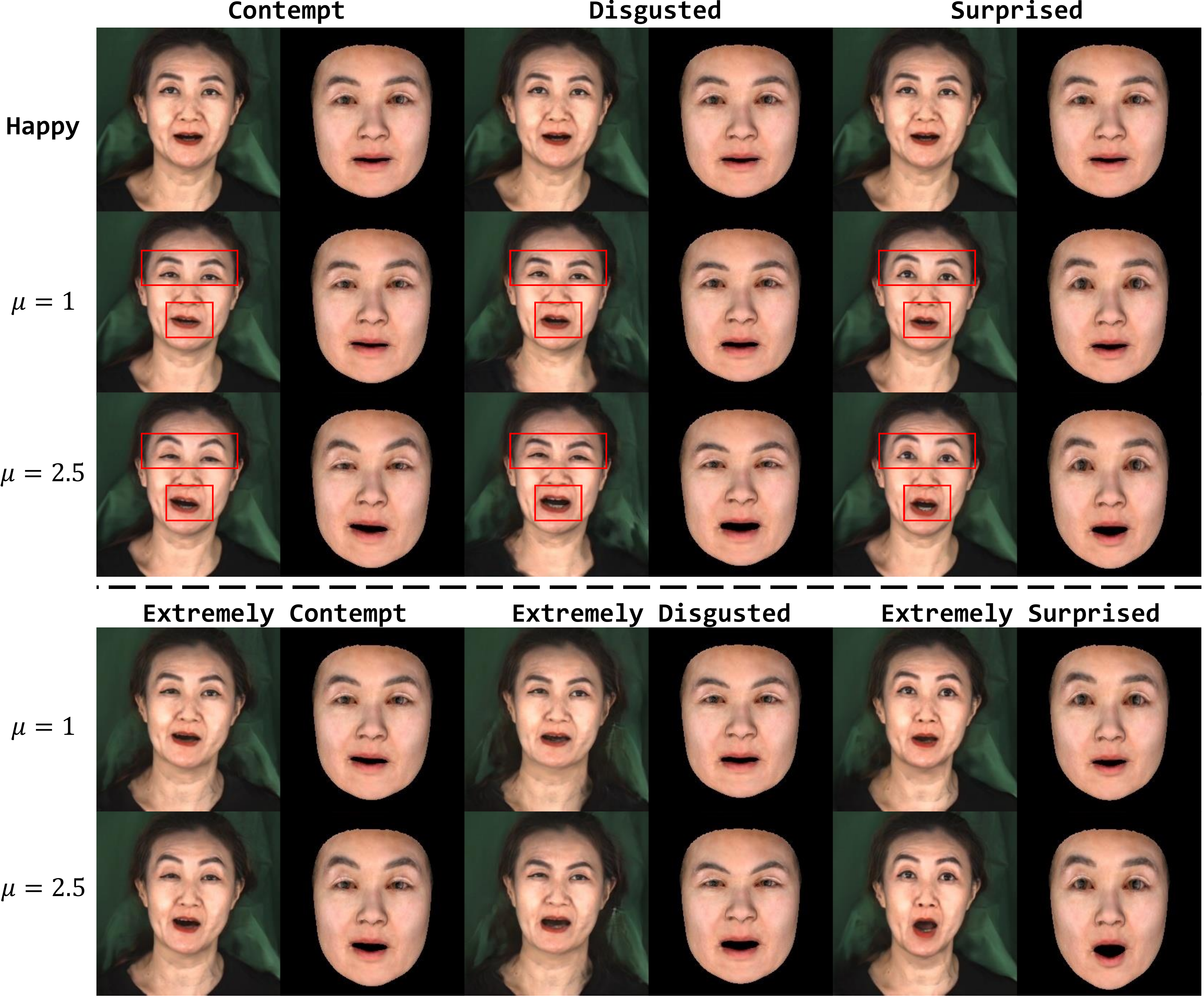}
	\caption{Results of different emotion styles and intensity levels. The top part shows the manipulation from the happy to three distinct emotion styles. The bottom part shows the results of style semantics that already encode intensity.
	}
	\label{fig:emo_control}
\end{figure}

% \subsection{Ablation Study and Further Analysis}
\noindent\textbf{Continuous Emotion Style Control.} We conduct a qualitative experiment to evaluate the effectiveness of our method for controlling emotion style. As shown in Fig.~\ref{fig:emo_control}, our method could change the emotion representation between two distinct styles, rather than previous methods only taking a neutral face as the source. We increase the intensity value from 1 to 2.5, which shows continuous and accurate expression changes. Please pay attention to the mouth and eyes regions. Furthermore, we explore the style semantics that already encode intensity, \emph{e.g.}, \textit{Extremely surprised}. Comparing rows 4, 5 with rows 2, 3, CLIP struggles to distinguish the intensity prompt. Thus the intensity token is essential in our method.

\begin{figure}
	\centering
	\includegraphics[width=0.38\textwidth]{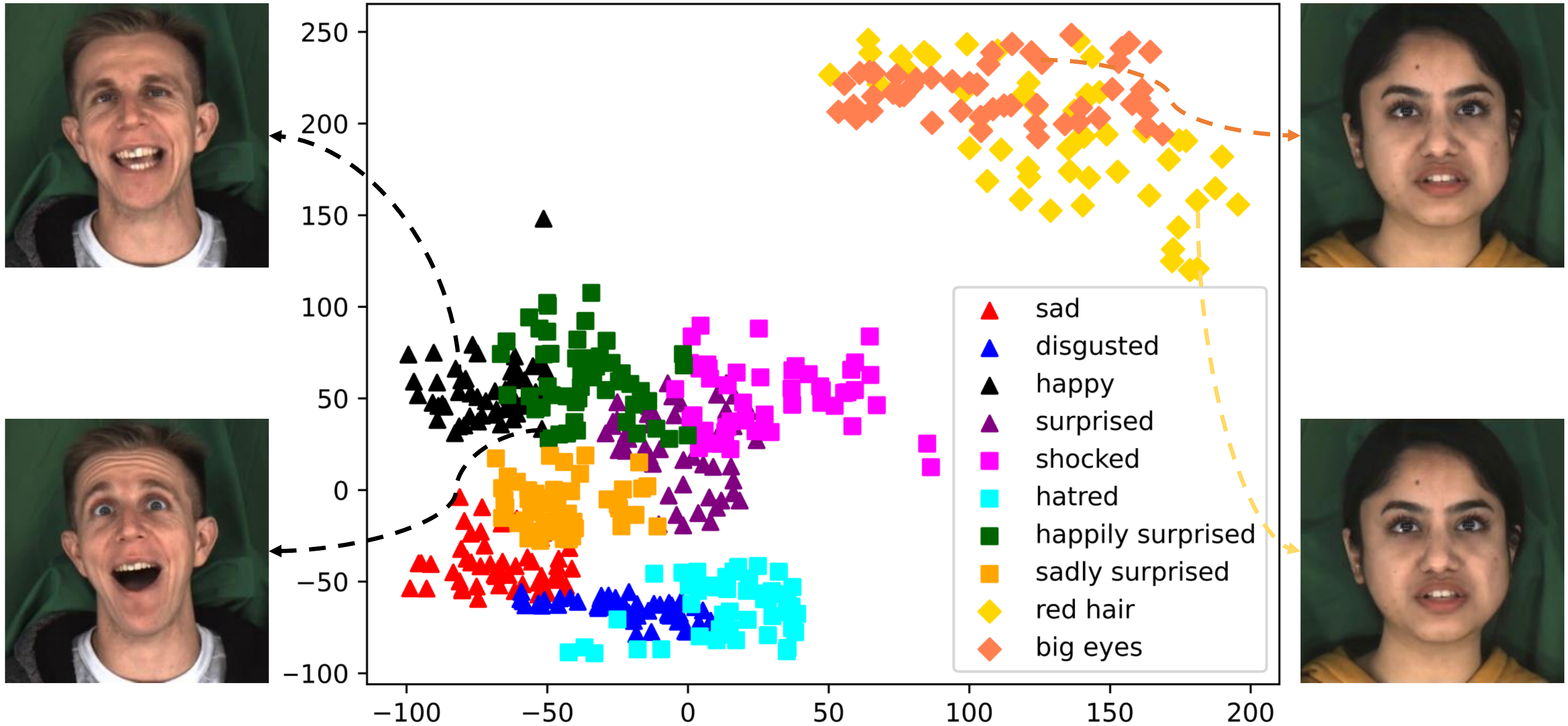}
	\caption{Clusters of the intensity token with the emotion and emotion-unrelated text descriptions. Markers $\triangle$, $\Box$, $\Diamond$ mean basic emotion, unseen emotion, and emotion-unrelated text prompts. 
	}
	\label{fig:tsne}
\end{figure}

% \noindent\textbf{Visualization of Attention Weights.} Fig.~\ref{fig:att} visualizes the average attention weights across all heads of the CLIP audio and Transformer encoders. We observe that the former focuses on some emotion-related timestamps to obtain precise style code, while the latter focuses on the local cues, which are more likely to influence the current expression. The relatively high weights of the first column in (b) indicate the prepended token effectively models the emotion intensity.

\noindent\textbf{Interpretability of Generalization.} To analyze the ability for unseen emotion generalization, we use t-SNE to visualize the latent codes of updated intensity token. As shown in Fig.~\ref{fig:tsne}, the four basic emotion styles (Marker $\triangle$) represent distinct clusters, while the clusters of those unseen emotion styles (Marker $\Box$) are mainly located in semantically similar areas. We further adopt two emotion-unrelated prompts (Marker $\Diamond$). Obviously, these two clusters are far from those of emotion and highly overlapped since both are meaningless to our model, \ie, their sampled faces are not changed in Fig.~\ref{fig:tsne}. Thus, our method could represent various styles located in similar emotion domains of CLIP.

\begin{figure}[t]
	\centering
	\includegraphics[width=0.45\textwidth]{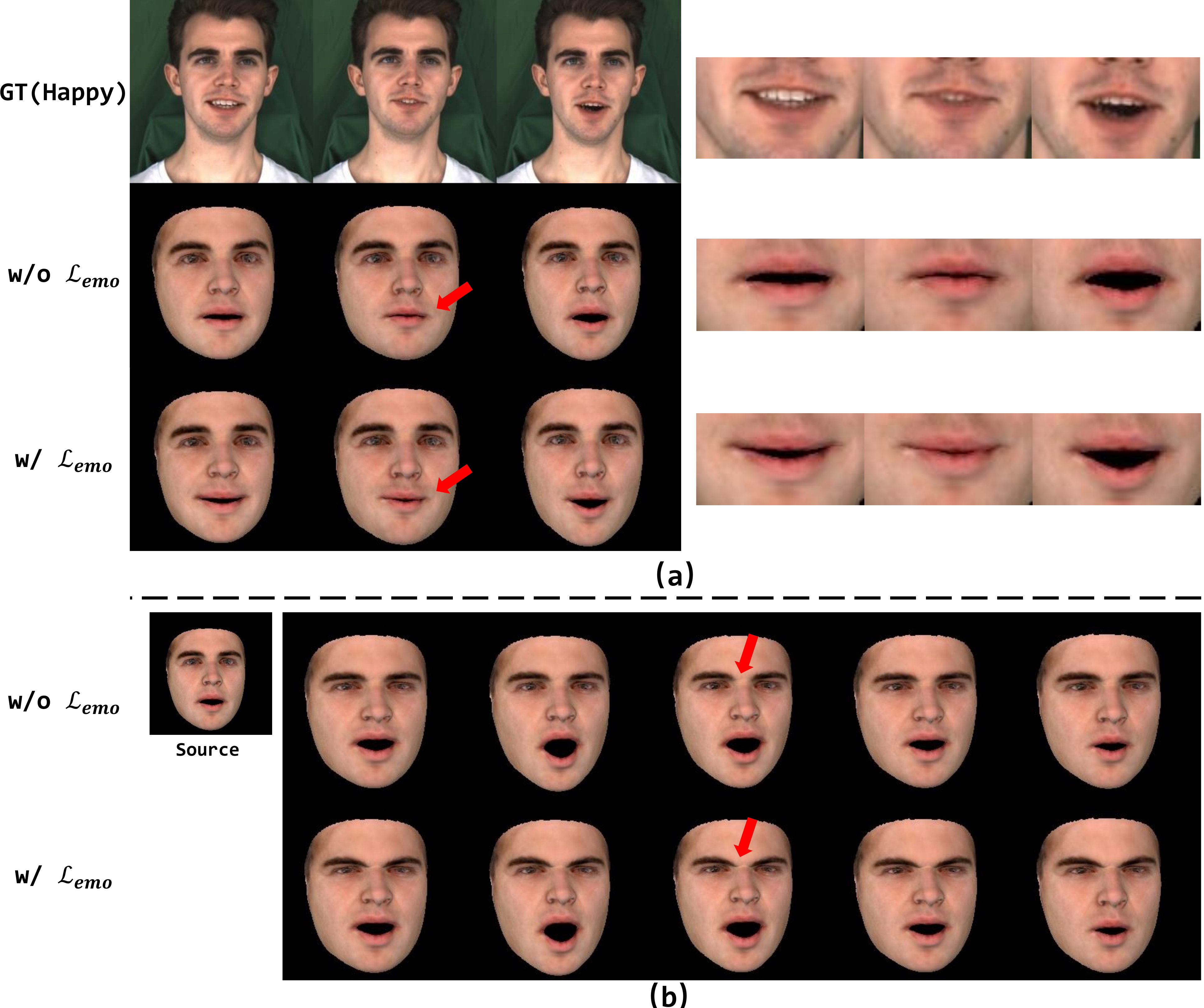}
	\caption{Qualitative ablation study for emotion consistency loss of EAC. (a) shows the effect on 3D face reconstruction of \textit{happy} emotion and (b) illustrates manipulation by \textit{angry} emotion. 
	}
	\label{fig:abla_loss}
\end{figure}

\subsection{Ablation Study and Efficiency Evaluation}

\noindent\textbf{Loss Functions.} Emotion consistency loss is critical to distill emotion cues from CLIP and improve the quality of expression coefficients. To verify its effectiveness, we present the qualitative results in Fig.~\ref{fig:abla_loss}. This loss term helps to capture precise and fine expression details, facilitating the 3D face reconstruction and emotion manipulation. Besides, consistent quantitative results are reported in Tab.~\ref{tab:ablation}.

\begin{figure}[t!]
	\centering
	\includegraphics[width=0.45\textwidth]{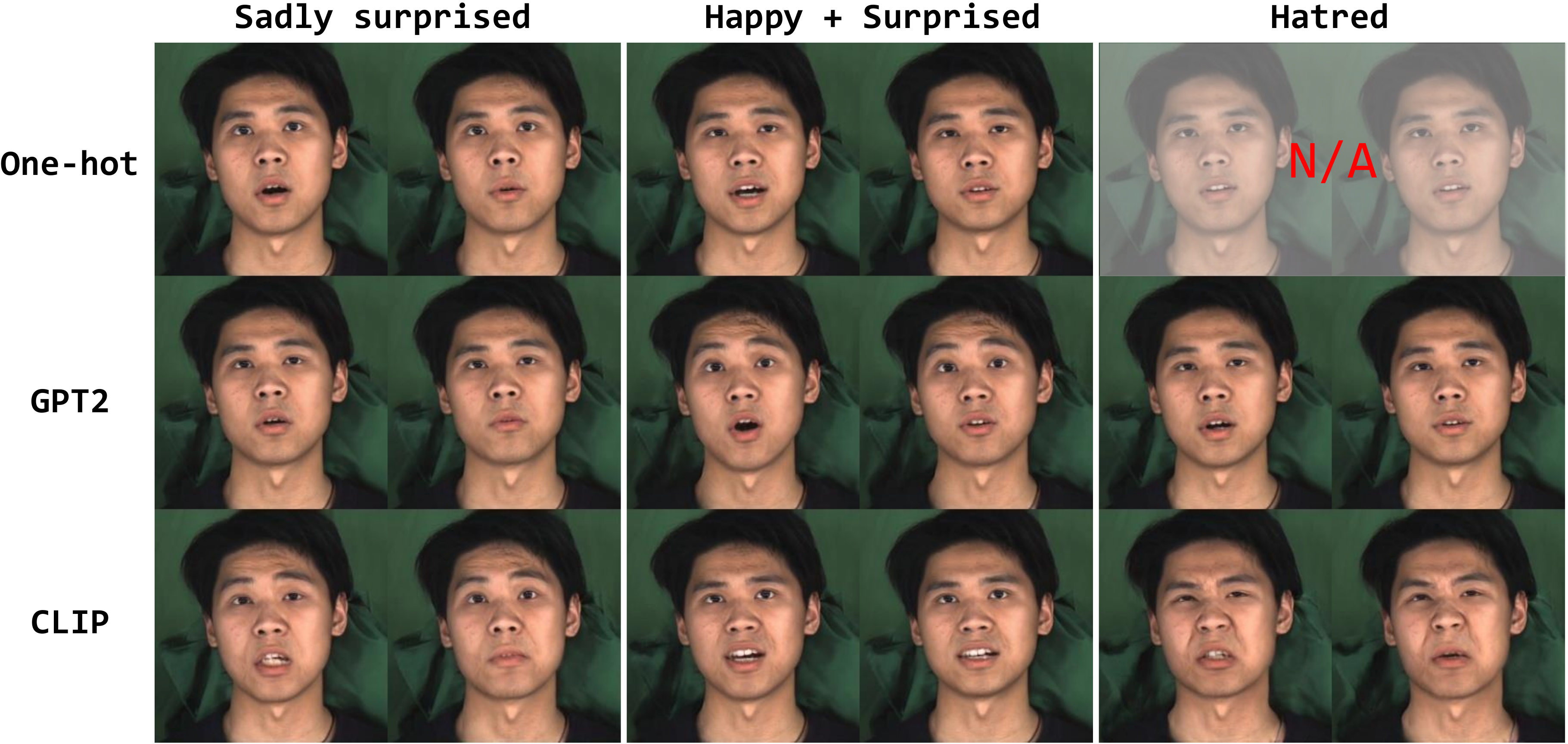}
	\caption{Qualitative ablation study of EAC with different emotion encodings on \textit{unseen styles}. This case is sampled from Fig.~\ref{fig:unseen_emo}. 
	}
	\label{fig:abla_model}
\end{figure}

\noindent\textbf{Emotion Encoding.} Tab.~\ref{tab:ablation} shows one-hot encoding and language pre-trained model GPT2~\cite{radford2019language} achieve comparable results with CLIP on pre-defined styles. We further explore the effect of these emotion encodings on the unseen emotion in Fig.~\ref{fig:abla_model}. One-hot fails to represent a new emotion style due to the fixed pattern, and compound styles due to lacking semantics. GPT2 is not available to the visual cues and struggles to reflect the unseen textual semantics to the image domain. Our method inherits rich visual and textual priors from CLIP, exhibiting better generalized ability.

\begin{table}[t!]
  \centering
  \scriptsize
   \renewcommand\arraystretch{1.0}
   \setlength\tabcolsep{6pt}
   \begin{tabular}{C{35pt}C{25pt}C{25pt}C{20pt}C{30pt}C{20pt}}
      \toprule
       \multicolumn{1}{c}{\multirow{2}{*}{Method}}  & \multicolumn{1}{c}{\multirow{2}{*}{Params (M)}} &\multicolumn{2}{c}{Training} &\multicolumn{2}{c}{Inference (1$\times$V100)} \\
       \cmidrule(lr){3-4} \cmidrule(lr){5-6}
        & &GPUs & days    & Memo (G)  & ms \\
      \midrule
      EAMM-256 & 101.89 & 4$\times$2080Ti & 5  & 3.8  & 21 \\
      Ours-256 & 62.38  & 4$\times$V100 & 3  & 2.4  & 26 \\
      Ours-512 & 62.97  & 8$\times$V100 & 7  & 3.2  & 37 \\
      \bottomrule
   \end{tabular}
   \caption{Efficiency evaluation during training and inference.}
   \label{tab:efficiency}
\end{table}

\noindent\textbf{Architecture of EAC.} To verify the effectiveness of the Transformer encoder in EAC, we replace it with stacked fully-connected layers or GRU-based recurrent neural networks. As shown in Tab.~\ref{tab:ablation}, our Transformer-based model obviously outperforms the above two architectures.

\begin{figure}[t!]
	\centering
	\includegraphics[width=0.4\textwidth]{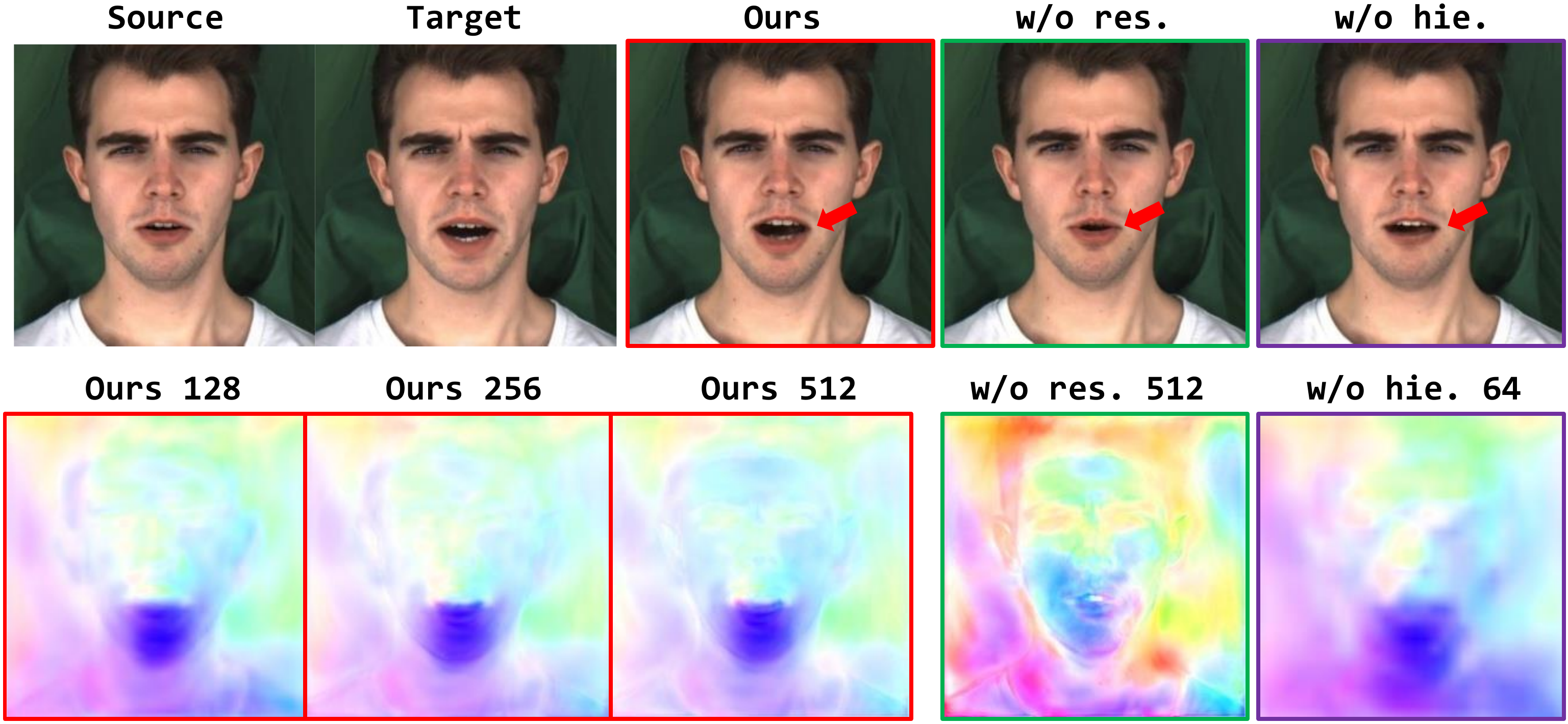}
	\caption{Qualitative ablation study of HEF with different flow estimation variants. We visualize the flow fields of w/o res. at scale 512 and the fixed $64 \times 64$ flow fields of w/o hie., both fail to model the precise movement, while our method gradually refines the high-resolution flow fields by hierarchical residual learning.
	}
	\label{fig:abla_flow}
\end{figure}

\noindent\textbf{Flow Estimation of HEF.} We further design two variants to explore the flow estimation structure, \ie, the one directly outputs the flow fields without residual refinement at each scale (w/o res.), and another one uses the fixed $64 \times 64$ flow field to adapt to the following high-resolution layers of HEF instead of further updating hierarchically (w/o hie.). Please refer to supplementary materials for modification details. The Fig.~\ref{fig:abla_flow} and Tab.~\ref{tab:ablation} verify the effectiveness of hierarchically learning deformation in the residual manner. Besides, we observe that the fixed low-resolution flow field cannot produce accurate animated high-resolution faces, which explains why FOMM~\cite{siarohin2019first} and EAMM are not competent for high-resolution generation.

\noindent\textbf{Efficiency Evaluation.} We report the running efficiency in the Tab~\ref{tab:efficiency}. The cost increases obviously as the resolution becomes larger. Notably, we achieve comparable training costs and inference speed with EAMM under the same resolution (256$\times$256), but our method is more \textit{memory-friendly}, \ie, lower model size and memory cost, which is compatible with relatively cheap devices, \eg, 1080Ti.

\begin{table}[t!]
  \centering
  \scriptsize
  \renewcommand\arraystretch{1.0}
  \setlength\tabcolsep{6pt}
  \begin{tabular}{C{40pt}C{30pt}C{30pt}C{30pt}}
      \toprule
      Method & EFD $\downarrow$ & LMD $\downarrow$ & Sync $\uparrow$ \\
      \midrule
      w/o $\mathcal{L}_{emo}$ & 0.096  & 2.40   & 3.53                 \\
      w/ $\mathcal{L}_{emo}$ & \textbf{0.065}   & \textbf{2.31}   & \textbf{3.57}               \\
      \midrule
      One-hot & 0.070  & 2.33    & 3.53     \\
      GPT2 & 0.067  & 2.33   & 3.56                 \\
      CLIP & \textbf{0.065}  & \textbf{2.31}    & \textbf{3.57}        \\
      \midrule
      MLPs & 0.122  & 3.54   & 2.23                 \\
      GRUs & 0.088   & 2.47   & 3.19               \\
      Transformers & \textbf{0.065}   & \textbf{2.31}   & \textbf{3.57}               \\
      \midrule
      w/o res. & 0.082  & 2.46   & 3.21                 \\
      w/o hie. & 0.076   & 2.42   & 3.25               \\
      Ours & \textbf{0.065}   & \textbf{2.31}   & \textbf{3.57}               \\
      \bottomrule
  \end{tabular}
  \caption{Quantitative ablation study with different losses and components, conducted on MEAD with \textit{basic styles} by default.}
  \vspace{-1.0em}
  \label{tab:ablation}
\end{table}

\section{Conclusions}
% In this paper, we present a flexible and practical emotional audio-driven face generation framework. Specifically, proposed EAC consists of a flexible CLIP-based emotion representation to distill the critical emotion information, and a Transformer-based structure converts the given audio clips and emotion cues to predict the expression coefficients. Then, a powerful texture generator HEF is proposed to integrate the style, appearance, and geometry information to obtain photo-realistic and emotion-aware faces. Our method could generalize to unseen emotion styles and identities without extra training. Qualitative and quantitative experiments demonstrate the superiority of our approach over other SOTA methods.

In this paper, we propose a novel one-shot emotional talking face generation framework. Specifically, a unified multi-modal CLIP-based emotion space and a texture generator are proposed to generalize to unseen emotions and guarantee the quality of animated faces, respectively. Qualitative and quantitative experiments demonstrate the superiority of our approach over SOTA methods.

\noindent\textbf{Acknowledgments.} This work is supported by the Key R\&D Program Project of Zhejiang Province (2021C01035).

% \noindent\textbf{Limitations.} Although EAC uses the Transformer to
% model temporal information, HEF is an image-based texture
% generator, which inevitably introduces temporal inconsistency.

%%%%%%%%% REFERENCES
{\small
\bibliographystyle{ieee_fullname}
\bibliography{main}
}

\end{document}